# Chatbot Application to Support Smart Agriculture in Thailand

Paweena Suebsombut
College of Arts, Media and Techology
and Institut Universitaire de
Technologie Lumière,
Chiang Mai University and Université
Lumière Lyon 2,
Chiang Mai (Thailand), Bron (France)
paweena_sueb@cmu.ac.th

Pradorn Sureephong
College of Arts, Media and Techology,
Chiang Mai University,
Chiang Mai, Thailand
dorn@camt.info

Aicha Sekhari
Institut Universitaire de Technologie
Lumière,
Université Lumière Lyon 2,
Bron, France
aicha.sekhari@univ-lyon2.fr

Suepphong Chernbumroong
College of Arts, Media and Techology,
Chiang Mai University
Chiang Mai, Thailand
suepphong@kic.camt.info

Abdelaziz Bouras
College of Engineering,
Qatar University,
Doha, Qatar
abdelaziz.bouras@qu.edu.qa

*Abstract*— A chatbot is a software developed to help reply to text or voice conversations automatically and quickly in real time. In the agriculture sector, the existing smart agriculture systems just use data from sensing and internet of things (IoT) technologies that exclude crop cultivation knowledge to support decision-making by farmers. To enhance this, the chatbot application can be an assistant to farmers to provide crop cultivation knowledge. Consequently, we propose the LINE chatbot application as an information and knowledge representation providing crop cultivation recommendations to farmers. It works with smart agriculture and recommendation systems. Our proposed LINE chatbot application consists of five main functions (start/stop menu, main page, drip irrigation page, mist irrigation page, and monitor page). Farmers will receive information for data monitoring to support their decision-making. Moreover, they can control the irrigation system via the LINE chatbot. Furthermore, farmers can ask questions relevant to the crop environment via a chat box. After implementing our proposed chatbot, farmers are very satisfied with the application, scoring a 96% satisfaction score. However, in terms of asking questions via chat box, this LINE chatbot application is a rule-based bot or script bot. Farmers have to type in the correct keywords as prescribed, otherwise they won't get a response from the chatbots. In the future, we will enhance the asking function of our LINE chatbot to be an intelligent bot.

## I. INTRODUCTION

"Smart Agriculture" is the concept of precision agriculture by using information and communication technology (ICT) to measure and control a crop and its environment, such as temperature, soil, etc., for optimum productivity. The main aim of smart agriculture is to use smart technologies to help farmers increase their income and agricultural sustainability, including resource wisely using smart technologies [1]. Therefore, adoption of smart internet of things (IoT) technologies in the agricultural domain helps farmers with production cost reduction and yield improvement. The major aim of IoT technology is to make it readable, recognizable, locatable, and addressable by using Wireless Sensor Networks (WSN), Radio Frequency Identification (RFID), or other means.

The IoT concept is adopted in various domains of the agriculture sector. For example, in the study [2], they literate the IoT application in precision agriculture. They proposed a monitoring system using wireless sensor networks and the internet. To provide data for agricultural research, they developed an information management system [2]. Meanwhile, "digital agriculture" based on IoT was proposed by Chen and Jin. The different sensor types are used to collect crop and field environment information, such as wind, soil water content, temperature, etc., and the information is analyzed by the system [3]. Furthermore, Wi-Fi based WSN in IoT applications is explained by Li Li et al., which is relevant to the IoT applications based on WSN, smart grid, and Wi-Fi [4]. The smart grid helps to improve the reliability of data collection, provide information accuracy, and provide an intelligent application of data collection. The intelligent application to monitor the environment based on IoT technology provides environmental information, such as air data, soil data, and water data, collected by using sensors and transferring the data to a cloud database or server for data processing and analyzing.

To summarize, the application of IoT technology provides numerous benefits to smart agriculture domains, affecting cost management and farm productivity. As mentioned earlier, several farming technologies have been implemented. However, the study [5] found that most of the local farmers in Chiang Mai, Thailand is aged and lack IT literacy, which makes it difficult to adopt new technologies. Thus, we propose that the chatbot could be the potential technology that would facilitate the user in the smart farming domain. Consequently, the aim of this paper is to propose a chatbot application to support farmers' decision-making for crop cultivation processes.

## II. LITERATURE AND RELATED WORKS

The Chatbot is an easy-to-use and very user-friendly application, especially for the elderly group [6]. Furthermore, the elderly are eager to communicate with this virtual agent because it has the ability to console the lonely elderly. Chatbot is one form of information system that is used to match an information source to a predefined or desired acknowledgement [7]. The chatbot application is an adoption of a computer program that is artificially intelligent (AI). It imitates the communication behaviors of humans, including spoken or text, using intelligent techniques such as video processing, Natural Language Processing (NLP) or image processing [8]. The Chatbot provides streamlined information, avoids repetitive tasks, helps to save time, answers difficult questions, and provides out-of-context information so that users can simplify their tasks. Thus, chatbot applications are developed to handle huge numbers of users in a cost-effective manner because they help to conquer the fallacious decisions and irrationalities that are caused by the behavior of humans.

"Conversational agents" is also another name for chatbots that have artificial intelligence backing, NLP, and machine learning (ML). The speech of a human being will be processed either in the form of voice or text to simulate an interaction or conversation with a real human. Users can use the chatbot via either standalone applications or web-based applications. Nowadays, chatbot applications are commonly used in several domains, which are mostly used in the customer service sector. The ability of a chatbot is to transfer data by searching for appropriate keywords for information collection in an easy way. A chatbot is a specific domain of conversational interface using a social network, chat solution, or messaging platform for communication. One of the AI techniques being used to design the chatbot is Pattern Matching [9]. With the design of an AI chatbot, we can create conversations and responses and actively learn by using NLP, which learns from conversations with customers. By using NLP techniques, a computer is able to understand the languages spoken by humans. For example, the application of NLP for grammar or spell checking, such as Google, Grammarly, etc. The predictive search typing of Google assists in suggesting the next word. Therefore, the command to remember the previous searched words helps to improve the response time and efficiency of the chatbot application [10].

### III. PROPOSED CHATBOT APPLICATION

There are several mobile chat applications used for communication. The survey result of [11] explains the highest technology usage of farmers in Thailand for communication (see Fig.1.).

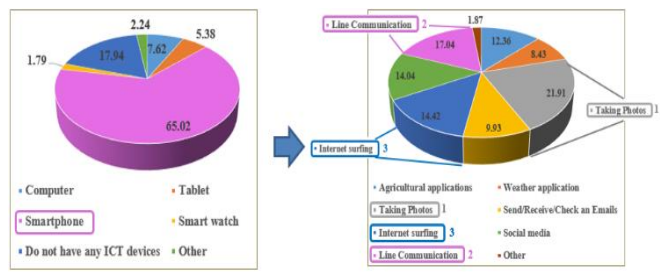

Fig. 1. Survey result of Thai farmer's technology usage
    (a) Technology device Usage
    (b) Reason of using technology device

Fig. 1. illustrates the result of the technology usage of Thai farmers. According to the findings, Thai farmers primarily use smartphones (65%) in their daily lives, as shown in Fig.1(a). In terms of communication platform, Thai farmers (17%) use the LINE chat mobile application to communicate with others, as shown in Fig.1(b). LINE is a freeware software application that allows users to communicate instantly on electronic devices. Such as cellphones, tablets, and desktop computers. LINE is the world's most popular mobile chat application, connecting users, friends, and family closer. Consequently, in this paper, we propose to use the LINE Chat mobile application as the information and knowledge representation for farmers to provide crop environment information, crop cultivation knowledge, and an appropriate recommendation to farmers.

Every day in the morning, our proposed LINE chatbot will send the knowledge video and current status to farmers. Moreover, during the day, it will send an appropriate recommendation for crop cultivation (irrigation, fertilization, disease control, insect pest control, and weed control) if there is some abnormal situation that harms the crop. Moreover, the proposed LINE chatbot supports data monitoring and smart irrigation control systems for farmers. Furthermore, farmers can ask a particular crop environment question via LINE chatbot. For this asking function, we use DialogFlow to set the keywords for asking, also known as the rule-based bot or script bot. The DialogFlow has a feature that helps users find the correct word when they type words. It will be used as the interface between Line chatbots and the system. However, we must interleave training sentences or words in order to train the DialogFlow.

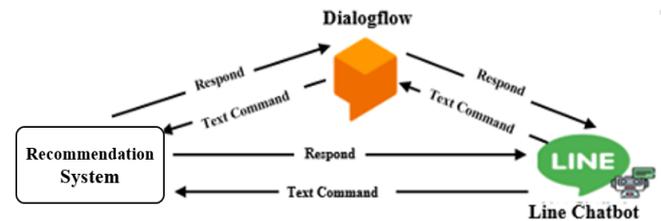

Fig. 2. LINE chatbot with DialogFlow and recommendation system architecture

Fig. 2 illustrates the architecture of our LINE chatbot with DialogFlow and a recommendation system. In terms of receiving recommendations, the recommendation system will automatically send the recommendation to the LINE chatbot when it finds some abnormal or harmful situation for the crop. In terms of data monitoring and control, the LINE chatbot will send the commands from farmers directly to the system. In terms of asking for some particular information, the LINE chatbot works with the DialogFlow to help farmers find the correct keywords, and then the DialogFlow will send the command to the system to get the requested information.

Fig.3 depicts flow of our proposed LINE chatbot with five primary functions:

- Start/Stop menu: farmers need to click 'Start' to begin using the LINE chatbot. If farmers encounter any problems, they can stop the LINE chatbot application by clicking on the 'START/STOP Menu.'

- Main page: is the interface to show all relevant information and recommendation to farmers both data monitoring and control system.

- Drip irrigation page: farmers can control 'ON/OFF' of the drip irrigation based on the recommendation.

- Mist irrigation page: farmers can control 'ON/OFF' of the mist irrigation system based on the recommendation.

- Monitor page: farmers will receive relevant information for crop cultivation for monitoring their

crop and field. Moreover, they can ask a particular question, such as "weather forecast," and the chatbot will respond with that information.

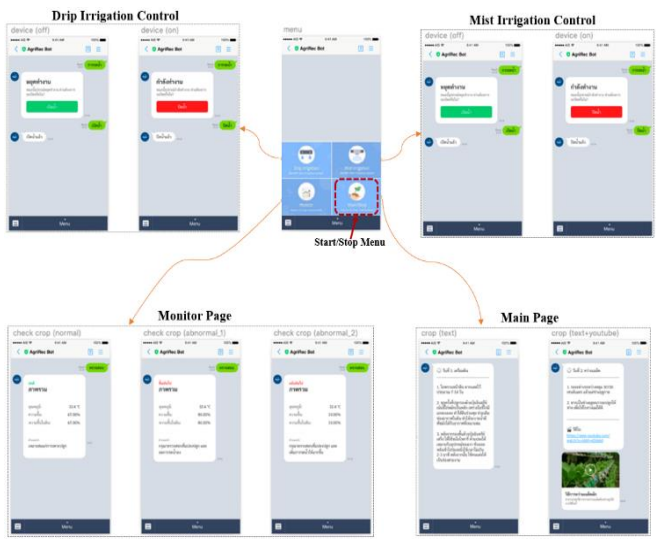

Fig. 3. Flow of proposed LINE chatbot application

## IV. IMPLEMENTATION RESULTS AND DISCUSSION

### A. Implementation Result

Our proposed LINE chatbot is implemented in Chiang Mai province (Thailand) for lettuce cultivation. Farmers added LINE friends by scanning the QR code. The LINE chatbot is implemented with smart agriculture technologies (sensors and IoT technology) to represent the knowledge and recommendations for crop cultivation processes including irrigation (drip and mist), fertilization, disease control, insect pest control, and weed control.

The LINE chatbot application automatically transfer the status of field information, knowledge, and an appropriate recommendation to farmers (see Fig.4 – Fig.7). Following receipt of the information and recommendations, the user may begin controlling the drip irrigation system to irrigate lettuce as directed. However, the user can always monitor their field via the 'MONITOR' page (see Fig.5). Furthermore, farmers can ask a particular question to get some information by typing text on the main page as shown in Fig.8.

After implementation, our proposed LINE chatbot application can support farmers for decision-making to cultivate crop. Farmers are very satisfied to our proposed LINE chatbot application with 96% of satisfaction score.

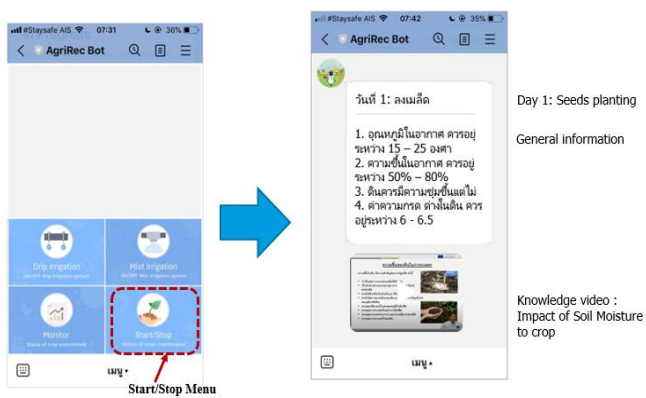

Fig. 4. Main page and Start menu of LINE chatbot for Lettuce cultivation

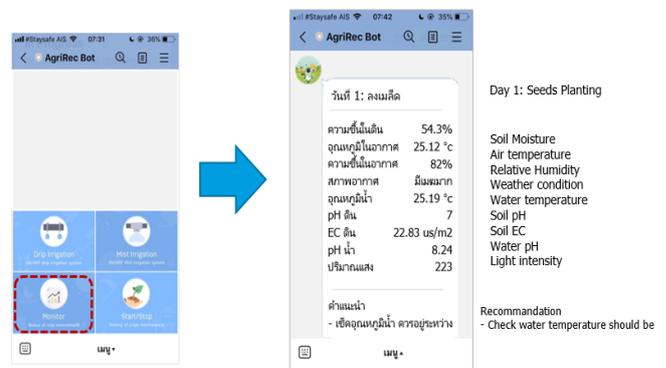

Fig. 5. Monitor page of LINE chatbot for Lettuce cultivation

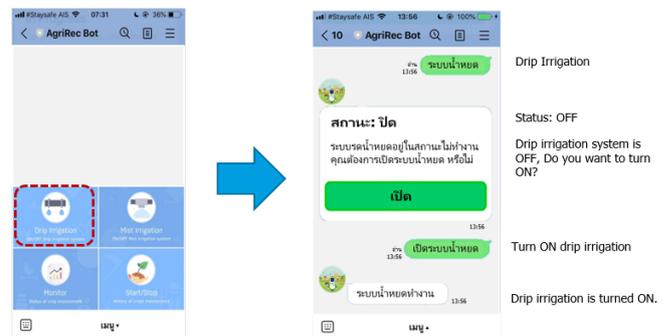

Fig. 6. Drip irrigation page of LINE chatbot for Lettuce cultivation

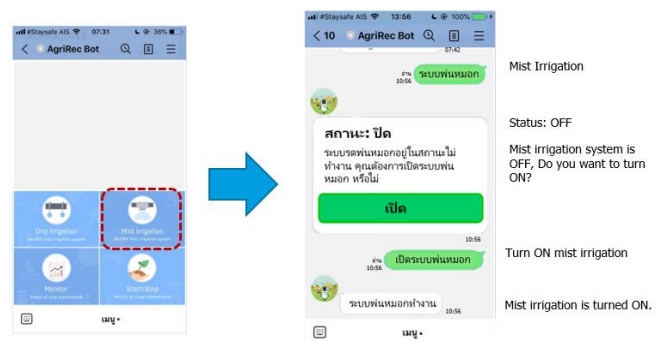

Fig. 7. Mist irrigation page of LINE chatbot for Lettuce cultivation

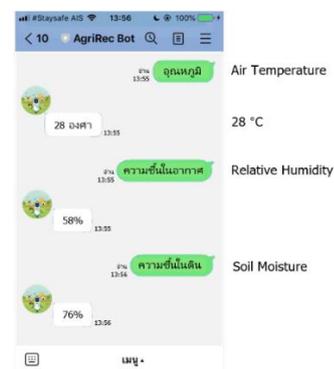

Fig. 8. Main page for asking a particular question of LINE chatbot for Lettuce cultivation

### B. Discussion

In order to get the information from the proposed LINE chatbot application, farmers must type the correct keywords to get the information, which is the limitation of this LINE chatbot application. However, the flexible function is required. This flexible function will help farmers get their requested information even if they type incorrect keywords. Therefore, our future work is the enhancement of the chat box function to help farmers receive their requested information even if they type incorrect keywords.

## V. CONCLUSION

This paper proposes the LINE chatbot mobile application to support farmers with crop cultivation processes (irrigation, fertilization, disease control, insect pest control, and weed control). The LINE chatbot will work with a recommendation system and IoT technologies to represent information, knowledge, and recommendations to farmers. Farmers can also use the LINE chatbot to monitor their crops and field environment, as well as control their irrigation systems (mist and drip irrigation). The LINE chatbot will receive the recommendation from the system automatically. Furthermore, farmers can ask for information via a chat box and DialogFlow function. However, farmers must type in the correct keywords to get the requested information. This is the limitation of the proposed LINE chatbot application that we plan to enhance in the future.

## ACKNOWLEDGMENT

The authors would like to acknowledge the support of the Institute Universitaire de Technologie (I.U.T) Lumière, University Lumière Lyon 2 (France), College of Arts, Media, and Technology, Chiang Mai University (Thailand), and the College of Engineering, Qatar University (Qatar). We would also like to acknowledge all of our colleagues who worked together and provided encouragement to the authors.

## REFERENCES

[1] J. Muangprathub, N. Boonnam, S. Kajornkasirat, N. Lekbangpong, A. Wanichsombat, and P. Nillaor, "IoT and agriculture data analysis for smart farm," Computers and Electronics in Agriculture, vol. 156, pp. 467-474, Jan. 2019.

[2] Zhao, Ji-chun, Jun-feng Zhang, Yu Feng, and Jian-xin Guo. "The study and application of the IOT technology in agriculture." In Computer Science and Information Technology ICCSIT, 2010 3rd IEEE International Conference on, vol. 2, pp. 462-465. IEEE, 2010.

[3] Chen, Xian-Yi, and Zhi-Gang Jin. "Research on key technology and applications for internet of things." Physics Procedia 33, pp. 561-566, 2011.

[4] Li, Li, Hu Xiaoguang, Chen Ke, and He Ketai. "The applications of WiFi-based wireless sensor network in internet of things and smart grid." In Industrial Electronics and Applications ICIEA, 2011 6th IEEE Conference on, pp. 789-793. IEEE, 2011

[5] P. Suebsombut, S. Chernbumroong, P. Sureephong, P. Jaroenwanit, P. Phuensane and A. Sekhari, "Comparison of Smart Agriculture Literacy of Farmers in Thailand," 2020 Joint International Conference on Digital Arts, Media and Technology with ECTI Northern Section Conference on Electrical, Electronics, Computer and Telecommunications Engineering (ECTI DAMT & NCON), 2020, pp. 242-245, doi: 10.1109/ECTIDAMTNCON48261.2020.90906

[6] W. Irene Yipei, Y. Ying, C. Shu-Jung, T. Wei -Ju, and T.-J. Sung, "Acceptance and sustainability of health promotion solutions for the elderly in Taiwan: Evidence from shi-lin elderly university in Taipei," in ACM International Conference Proceeding Series, 2019, pp. 21–27.

[7] J. Trivedi, "Examining the Customer Experience of Using Banking Chatbots and Its Impact on Brand Love: The Moderating Role of Perceived Risk," J. Internet Commer., vol. 18, no. 1, pp. 91–111, 2019.

[8] K. Patil and M. S. Kulkarni, "Artificial intelligence in financial services: Customer chatbot advisor adoption," Int. J. Innov. Technol. Explor. Eng., vol. 9, no. 1, pp. 4296–4303, 2019.

[9] M. Dahiya, "A Tool of Conversation: Chatbot, International Journal of Computer Sciences and Engineering, Volume-5, Issue-5 E-ISSN: 2347-2693," Int. J. Comput. Sci. Eng., vol. 5, no. December, 2017.

[10] V. Aggarwal, A. Jain, H. Khatter, and K. Gupta, "Evolution of Chatbots for Smart Assistance," no. 10, pp. 77–83, 2019.

[11] Suebsombut P, Chernbumroong S, Sureephong P, Jaroenwanit P, Phuensane P, Sekhari A. Comparison of Smart Agriculture Literacy of Farmers in Thailand. In 2020 Joint International Conference on Digital Arts, Media and Technology with ECTI Northern Section Conference on Electrical, Electronics, Computer and Telecommunications Engineering (ECTI DAMT & NCON) 2020 Mar 11 (pp. 242-245). IEEE.